\documentclass[twoside,11pt]{article}

\usepackage{jmlr2e}
\usepackage{amsmath}
\usepackage{mdframed,lipsum}
\usepackage{multirow}
\usepackage{color}
\usepackage{url}
\usepackage{tikz}
\usepackage{booktabs}
\usepackage{bbm, dsfont}
\usepackage{algorithm,algorithmic}

\DeclareMathOperator*{\argmin}{arg\,min}
\newcommand{\nonl}{\renewcommand{\nl}{\let\nl\oldnl}}
\newcommand{\fin}{f_\iota}
\newcommand{\Di}{\mathcal{D}_\iota}
\newcommand{\Db}{\mathcal{D}_b}
\newcommand{\fb}{f_b}
\newcommand{\mD}{\mathcal{D}}
\newcommand{\mR}{\mathcal{R}}

\jmlrheading{1}{2019}{1-48}{4/00}{10/00}{meila00a}{Tong Wang and Qihang Lin}

\ShortHeadings{Hybrid Predictive Model}{Tong Wang, Qihang Lin}
\firstpageno{1}
\begin{document}
    
    \title{Hybrid Predictive Model: When an Interpretable Model Collaborates with a Black-box Model
    }
    
    \author{\name Tong Wang  \email tong-wang@uiowa.edu \\
        \addr Tippie College of Business\\University of Iowa \\ Iowa City, IA, USA
        \AND \name Qihang Lin  \email qihang-lin@uiowa.edu \\
        \addr  Tippie College of Business\\University of Iowa  \\ Iowa City, IA, USA
    }

    \editor{}
    \maketitle

    \begin{abstract}
Interpretable machine learning has become a strong competitor for traditional black-box models. However, the possible loss of the predictive performance for gaining interpretability is often inevitable, putting practitioners in a dilemma of choosing between high accuracy (black-box models) and interpretability (interpretable models).  In this work, we propose a novel framework for building a Hybrid Predictive Model (HPM) that integrates an interpretable model with \textbf{\emph{any}} black-box model to combine their strengths. The interpretable model substitutes the black-box model on a subset of data where the black-box is overkill or nearly overkill, gaining transparency at no or low cost of the predictive accuracy. We design a principled objective function that considers predictive accuracy, model interpretability, and model \emph{transparency} (defined as the percentage of data processed by the interpretable substitute.) Under this framework, we propose two hybrid models, one substituting with association rules and the other with linear models, and we design customized training algorithms for both models. We test the hybrid models on structured data and text data where interpretable models collaborate with various state-of-the-art black-box models. Results show that hybrid models obtain an efficient trade-off between transparency and predictive performance, characterized by our proposed \emph{efficient frontiers}.

    \end{abstract}
    
\section{Introduction}
As data from various fields have seen rapid growths in volume, variety and velocity \citep{zikopoulos2012understanding}, machine learning models tend to grow bigger and more complicated in structure, especially with the recent break out of deep learning. 
The predictive performance, however, is not obtained for free. 
Systems such as deep neural networks and ensemble models are \emph{black-box} in nature, which means the models have a complex and opaque decision-making process that is difficult for humans to understand. In many heavily regulated industries such
as judiciaries and healthcare, black-box models, despite their good predictive performance, often find it challenging to find their way into real-world deployment since domain experts are often reluctant to trust and adopt a model if they do not understand it. EU's General
Data Protection Regulation (GDPR) has recently called for the  ``right to explanation'' (a right to information about individual decisions made by algorithms)  \citep{parliament,phillips2018international} that requires human understandable predictive processes of models.

To facilitate human understandability, interpretable machine learning is proposed.  Interpretable machine learning models use a small number of cognitive chunks presented in a human-understandable way (e.g., linear, logical, etc.) to construct a model so users can comprehend how predictions are produced. Various forms of interpretable models have been developed, including rule-based models \citep{wang2017bayesian,lakkaraju2017interpretable}, sparse linear models \citep{zeng2017interpretable}, case-based reasoning \citep{richter2016case}, etc. 
They are easy to understand, use, and diagnose compared to complex black-box models. 

Yet, the benefits of interpretable machine learning models often come at a price of compromised predictive performance. While the state-of-the-art research in interpretable machine learning has further pushed the limit of the models, and they can, sometimes, perform as well as black-box models, it is more often observed that interpretable models lose to black-box models on complicated tasks  since they need to have a simple structure to maintain interpretability while black-box models are trained to optimize a single objective, the predictive performance. In this paper, we focus on this more common situation where the best model is a black-box and choosing any of its interpretable competitors will lead to a loss in predictive performance.

A popular solution to still gain interpretability in the above situation is to use black-box explainers, which generate post hoc explanations for a black-box model, either locally \citep{ribeiro2016should} or globally \citep{adler2016auditing,lakkaraju2017interpretable}. However, two concerning issues exist.  First, explainers only approximate but do not characterize exactly the decision-making process of a black-box model, yielding an imperfect explanation fidelity.
Second, ambiguity and inconsistency \citep{ross2017right,lissack2016dealing} exist in explanations since different explanations can be generated for the same prediction, by different explainers or the same explainer with different parameters. Both issues result from the fact that explainers are not the decision-making process themselves and only approximate it afterward.

In this paper, we propose an alternative solution that introduces interpretability in a novel way. Our solution does not aim to explain a black-box but to substitute it, on a subset of data where the black-box model is overkill or nearly overkill.
We argue that even if a complex black-box model has the best predictive performance overall, it is not necessarily the best \emph{everywhere} in the data space. We hypothesize that 
\emph{there exists a subspace in the feature space where a simpler (interpretable) model can be as accurate as or almost as accurate as the black-box model}. Therefore, the black-box model could be replaced by the simple model with no or low cost of the predictive performance while gaining interpretability.

If such an interpretable substitute can be found, we can design a hybrid predictive model where input is first sent to the interpretable model to see if a prediction can be directly generated, if not, the black-box model will be activated. We call the percentage of data processed by an interpretable model \textbf{transparency} of the model, which represents how often a prediction is interpretable. The goal of a hybrid model is to ``squeeze'' transparency from the decision-making process at a minimum cost of predictive performance. 

To build a hybrid model, we need to address the following research questions. How to find such a sub-area in the feature space? What kind of interpretable models can be used as a local substitute for a complex model? And what is the form of the decision-making process? Finally, what does a system benefit from utilizing an interpretable local substitute?

To answer these question, we design a general framework for hybrid predictive models and formulate a principled objective that considers predictive performance, complexity of the interpretable substitute and the transparency of the model. We instantiate this framework with the two most popular forms of interpretable models in this paper, association rules and linear models, to build two hybrid models with customized training algorithms.  We apply both models to structured datasets and text data, where the interpretable models partially substitute state-of-the-art black-box models including ensembles and neural networks. To evaluate the models, we propose \textbf{efficient frontiers} that characterize the trade-off between transparency and accuracy. Efficient frontiers unify interpretable models, black-box models, and hybrid models, with black-box models located at transparency equal to zero, interpretable models at transparency equal to one, and hybrid models connecting the two extremes and spanning the entire spectrum of transparency. 

The proposed framework forms a collaboration between an interpretable model and a black-box model, utilizing the strengths of them where they are most competent. An important advantage of the proposed framework is that it is \textbf{agnostic} to the black-box model, which means any implementation or algorithmic detail of the black-box model or even what kind of model it is remains unknown during training. We only need the input and output of the black-box and the true labels of the input to train a hybrid model. This training mechanism will largely conceal information of the black-box system from its collaborator. This property is critical in building collaborations among different systems which do not wish to share technical details.

Finally, we would like to emphasize that the proposed model is not a black-box explainer, but can serve as a pre-step for a black-box explainer: we first find a region that can be taken over by an interpretable model and then sends the rest of the data to a black-box to predict and an explainer to explain.



The rest of the paper is organized as follows. Section \ref{sec:literature} reviews related work on interpretable machine learning. We present the framework of hybrid decision making in Section \ref{sec:hybrid} and instantiate the model with association rules and linear models in Section \ref{sec:rules} and Section \ref{sec:linear}, respectively, where we cast the two interpretable models into the hybrid framework and design customized training algorithms. In Section \ref{sec:experiments}, we test the performance of the hybrid models on structured and text datasets from different domains where interpretability is most pursued and discuss the results in detail. 
\section{Literature Review}\label{sec:literature}
Our work is broadly related to new methods for interpretable machine learning, most of which fall into two categories.  
The first is developing models that are interpretable stand-alone and do not need external explainers. These interpretable models often have a few cognitive chunks and simple logic to be easy to understand. Previous works in this category include rule-based models such as rule sets \citep{wang2017bayesian, rijnbeek2010finding,mccormick2011hierarchical}, rule lists \citep{yang2017scalable,angelino2017learning}. 
 Rule-based models check an input against a set of ordered or unordered rules. If an input is captured by a rule, the corresponding consequence of the rule is produced as the prediction. Linear models, especially sparse linear systems \citep{zeng2017interpretable, ustun2016supersparse, koh2015two} are also a canonical and popular form of interpretable models. They learn a possibly small set of non-zero weights for features and then compare the weighted sum of features with a threshold to determine the class labels.  When learning an interpretable model, the model complexity is regularized to improve interpretability and avoid overfitting. For rule-based models, model regularization often refers to reducing the number of rules, the number of features, the total number of conditions, etc \citep{wang2018multi, lakkaraju2016interpretable}. Sparse linear models regularize the number of non-zero coefficients or force the coefficients to be integers \citep{ustun2016supersparse}. We use rules and linear models to build two types of hybrid models in this paper.

The second line of research in interpretable machine learning is to provide explanations for a black-box model, locally \citep{ribeiro2016should} or globally \citep{adler2016auditing, lakkaraju2017interpretable}, providing some insights into the black-box model by identifying key features, interactions of features \citep{tsang2017detecting}, etc. 
One representative work is LIME \citep{ribeiro2016should} that explains the predictions of any classifier in an interpretable and faithful manner, by learning an interpretable linear model locally around the prediction. Others like \cite{ribeiro2018anchors, lakkaraju2019faithful} use association rules to explain a black-box system. More recently, developments in deep learning have been connected strongly with interpretable machine learning \citep{goodfellow2014explaining,frosst2017distilling} and have
contributed novel insights into representational issues.  But there has been recent debate \citep{rudin2018please} on potential issues of black-box explainers since explainers only approximate but do not characterize exactly the decision-making process of a black-box model, often yielding an imperfect fidelity. In addition, there exists ambiguity and inconsistency \citep{ross2017right,lissack2016dealing} in the explanation since there could be different explanations for the same prediction generated by different explainers, or by the same explainer with different parameters. Both issues result from the fact that the explainers only approximate in a post hoc way. They are not the decision-making process themselves.

Our work is fundamentally different from the research above. A hybrid predictive model is not a pure interpretable model. It utilizes the partial predictive power of black-box models to preserve the predictive accuracy. It is also not an explainer model which only observes but does not participate in the decision process. Here, a hybrid predictive model uses an interpretable model and a black-box model jointly to make a prediction.  

There exist a few singleton works that combine multiple models. One of the earliest works \citep{kohavi1996scaling} proposed NBTree which combines decision-tree classifiers and Naive-Bayes classifiers, \citet{shin2000hybrid} proposed a system combining neural network and memory-based learning. \citet{hua2006hybrid} combined SVM and logistic regression to forecast intermittent demand of spare parts, etc. A recent work \citep{wang2015trading} divides feature spaces into regions with sparse oblique tree splitting and assign local sparse additive experts to individual regions. Aside from the singletons, there has been a large body of continuous work on neural-symbolic or neural-expert systems \citep{garcez2015neural} pursued by a relatively small research community over the last two decades and has yielded several significant results \citep{mcgarry1999hybrid,garcez2012neural,taha1999symbolic,towell1994knowledge}.   This line of research has been carried on to combine deep neural networks with expert systems to improve predictive performance \citep{hu2016harnessing}. 

Compared to the collaborative models discussed above, our model is distinct in that the proposed framework can work with \textbf{\emph{any}} black-box model. This black-box model could be a carefully calibrated, advanced model using confidential features or techniques. Our model only needs predictions from the black-box model and do not need to alter the black-box during training or know any other information from it. This minimal requirement of information from the black-box collaborator renders much more flexibility in creating collaboration between different models, largely preserving confidential information from the more advanced partner.

\section{Hybrid Predictive Model}\label{sec:hybrid}
We present a general framework for building a hybrid predictive model and formulate an objective function combining important properties of a predictive model.  
  Let $\mathcal{D} = \{(\mathbf{x}^{(n)},y^{(n)})\}_{n=1}^N$  represent a set of training examples where $\mathbf{x}^{(n)}\in\mathcal{X}$ is a set of attributes for instance $n$ and  $y^{(n)}\in \mathcal{Y}$ is the target variable. In this paper we focus on binary classification so $y^{(n)} \in \{-1,1\}$. Let there be a black-box model $\fb$ which can be any pre-trained model with unknown type.  We are only provided with its predictions on the training set, denoted as $\{\widehat{y_b}^{(n)}\}_{n=1}^N$. 
  
\begin{figure}[h!]
\centering
  \includegraphics[width=0.6\textwidth]{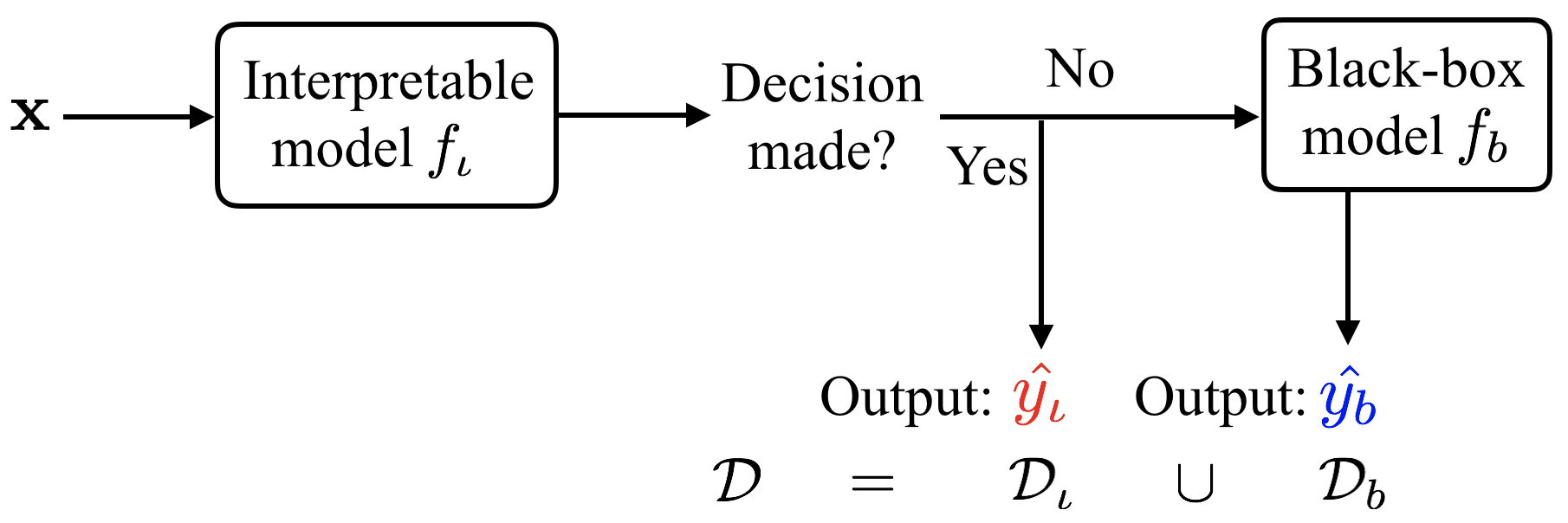}
\caption{A decision-making process of a hybrid predictive model.}\label{fig:hybrid}    
\end{figure}

  Given $\{(\mathbf{x}^{(n)},y^{(n)}, \widehat{y_b}^{(n)})\}_{n=1}^N$, our goal is to build an interpretable model $\fin$ that replaces $\fb$ on a subset of data, forming a hybrid predictive model, denoted as $f=\langle    \fin,\fb\rangle$.  Let $\Di$ represent the subset of data processed by $\fin$ and $\Db$ is the remaining data processed by $\fb$. 
  We design the predictive process as below: an input $\mathbf{x}$ is first sent to $\fin$. If a prediction can be generated, $\fin$ will directly output a prediction $\widehat{y}_\iota$, else, $\mathbf{x}$ is sent to $\fb$ to generate a prediction $\widehat{y}_b$.  See Figure~\ref{fig:hybrid} for an illustration.

We consider three properties critical when building a hybrid predictive model. \emph{First}, we desire a model with high \textbf{predictive performance}. Since $\fb$ is pre-given, the accuracy of a hybrid model is determined by two factors, the predictive accuracy of $\fin$ and the collaboration of $\fin$ and $\fb$, i.e., the partition of $\mathcal{D}$ into $\Di$ and $\Db$. $\fin$ and $\fb$ being completely different models allows the hybrid model to exploit the strengths of the two if the training examples are partitioned strategically, sending examples to the model which can obtain the maximum benefit. \emph{Second}, we desire high \textbf{model interpretability} of $\fin$. Bringing interpretability into the decision-making process is one of the motivations of building a hybrid model. The definition of interpretability is model specific and usually refers to having a small size and low model complexity, often measured by the number of cognitive chunks \citep{doshi2017roadmap}. For example, in rule-based models, the smallest cognitive chunks are conditions (feature-value pairs) in rules. The total number of conditions in a model represents the complexity of the model \citep{wang2018multi}. For linear models, the smallest cognitive chunks are features, and therefore sparse linear models regularize the number of non-zero coefficients \citep{ning2011slim}.
\emph{Finally}, we want to maximize the amount of data sent to $\fin$, which represents how often a prediction is interpretable, i.e., captured by $\fin$. To do that we define a novel metric, \textbf{transparency}, which is the percentage of $\Di$ in $\mathcal{D}$. 
\begin{definition}\label{def:exp}
The transparency of a hybrid model $f = \langle \fin,\fb \rangle$ on $\mathcal{D}$ is the percentage of data processed by $\fin$, i.e., $\frac{\Di}{\mathcal{D}}$, denoted as $\mathcal{E}(f,\mathcal{D})$.    
\end{definition}
 Following this definition, a stand-alone interpretable model has transparency equal to one, and a black-box model has transparency equal to zero.

In light of the  considerations above, we formulate a learning objective $\Lambda(f_\iota,\mathcal{D},\{\widehat{y}_b^{(n)}\}_{n=1}^N)$ as a linear combination of the three metrics. 
Given training data $\mathcal{D}$,  predictions of a pre-trained black-box on $\mathcal{D}$, $\{\widehat{y}_b^{(n)}\}_{n=1}^N$, and a pre-defined form of interpretable models $\mathcal{F}_\iota$, our goal is to find the optimal $f_\iota^*$ that
\begin{equation}\label{eqn:obje}
    f_\iota^* \in \arg\min_{f_\iota \in \mathcal{F}_\iota} \Lambda(f_\iota,\mathcal{D},\{\widehat{y}_b^{(n)}\}_{n=1}^N).
\end{equation}

Next, we instantiate the proposed framework with selected interpretable substitutes. An interpretable substitute needs to fulfill two objectives, to produce a human-understandable prediction and to partition the data.  We choose two interpretable models as examples, association rules and linear models, corresponding to the model space of $\mathcal{F}_\text{rule}$ and $\mathcal{F}_\text{linear}$, respectively. Both are popular forms of models in the interpretable machine learning literature. In the next two sections, we will cast the two interpretable models into the proposed framework and develop customized training algorithms optimizing the objective (\ref{eqn:obje}).

\section{Hybrid \emph{Rule Set}}\label{sec:rules}
In this section, we choose association rules for $f_\iota$ and call the model Hybrid Rule Set. We use $\mathcal{F}_\text{rule}$ to denote the model space for $\fin$. Rules are easy to understand for their simple logic and symbolic presentation. They also naturally handle the partition of data by separating examples according to if they satisfy the rules. 

\subsection{Model Formulation}
We construct two rule sets for $\fin$, a \emph{positive rule set}, $\mathcal{R}_+$, and a \emph{negative rule set}, $\mathcal{R}_-$.  If $\mathbf{x}^{(n)}$  satisfies any rules in $\mathcal{R}_+$, it is classified as positive. Otherwise, if it satisfies any  rules in $\mathcal{R}_-$, it is classified as negative. A decision produced from rules is denoted as $\widehat{y_\iota}^{(n)}$. If $\mathbf{x}^{(n)}$ does not satisfy any rules in $\mathcal{R}_+$ or $\mathcal{R}_-$, it means $\fin$ fails to decide on $\mathbf{x}^{(n)}$. Then $\mathbf{x}^{(n)}$ is sent to the black-box model $f_b$ to generate a decision $\widehat{y_b}^{(n)}$. The predictive process of $f$ is  shown below.
\begin{align}
&\textbf{if } \mathbf{x}^{(n)} \text{ obeys } \mR_+, Y = 1 \notag \\
&\textbf{else if }\mathbf{x}^{(n)} \text{ obeys } \mR_-, Y = -1 \notag \\
&\textbf{else }Y= \fb(\mathbf{x})
\end{align}
Here $\mR_+$ and $\mR_-$ are embedded in an ``else if '' logic. If an instance satisfies both $\mR_+$ and $\mR_-$, it is classified as positive since $\mR_+$ is ranked higher.  Define $\mathcal{R} = (\mathcal{R}_+,\mathcal{R}_-)$. We will use $\mR$ and $\fin$ interchageably for hybrid rule sets.

The association rules not only produce predictions but also naturally partition the feature space into $\Di$ and $\Db$, covering a subset of ``certainly positive'' instances with $\mR_+$ and a subset of ``certainly negative'' instances with $\mR_-$.

 We define the coverage of rules below.
\begin{definition}
A rule $r$ \emph{covers} $\mathbf{x}^{(n)}$ if $\mathbf{x}^{(n)}$ obeys the rule, denoted as $\text{covers}(r,\mathbf{x}^{(n)}) = 1$.
\end{definition}
\begin{definition}
A set of rules $R$ \emph{covers} $\mathbf{x}^{(n)}$ if  $\mathbf{x}^{(n)}$ obeys at least one rule in $R$, i.e. 
\begin{equation}
\text{covers}(R,\mathbf{x}^{(n)}) = \mathbbm{1}\left(\sum_{r\in R}\text{covers}(r,\mathbf{x}^{(n)})\geq 1\right).
\end{equation}
\end{definition}
\begin{definition}\label{def:supp}
Given a data set $\mathcal{D}$, the \textit{support} of a rule set $\mR$ in $\mathcal{D}$ is the number of observations covered by $R$, i.e., 
\begin{equation}
    \text{support}(\mathcal{\mR}, \mD) = \sum_{n=1}\text{covers}(\mR,\mathbf{x}^{(n)}).
\end{equation}
\end{definition}

\vspace{-2mm}
\subsubsection*{Learning Objective}
Now we cast learning objective $\Lambda(f_\iota,\mathcal{D},\{\widehat{y}_b^{(n)}\}_{n=1}^N)$ to Hybrid Rule Sets. 
First, we measure the predictive performance with misclassification error, represented $\ell(\langle\mathcal{R},\fb\rangle,\mathcal{D})$.
\begin{align}
&\ell(\langle\mathcal{R},\fb\rangle,\mathcal{D})= \sum_{n=1}^N \bigg(\frac{(1-y^{(n)})}{2}\text{covers}(\mathcal{R}_+, \mathbf{x}^{(n)})\!\!\! &&\!\!\triangleright\text{errors from }\mathcal{R}_+ \notag \\
&+\frac{(1+y^{(n)})}{2}\Big(1-\text{covers}(\mathcal{R}_+, \mathbf{x}^{(n)})\Big)\text{covers}(\mathcal{R}_-,\mathbf{x}^{(n)})\!\!\!  &&\!\!\triangleright      \text{errors from }\mathcal{R}_- \notag \\
&+\big((1-\text{covers}(\mathcal{R}_+, \mathbf{x}^{(n)})(1-\text{covers}(\mathcal{R}_-, \mathbf{x}^{(n)}) \big) \!\!\!&& \!\!\triangleright\text{instances not covered by $\mathcal{R}_+$ or $\mathcal{R}_-$} \notag \\
&\times\big(\frac{(1+y^{(n)})}{2}\frac{(1-\widehat{y_b}^{(n)})}{2} + \frac{(1-y^{(n)})}{2}\frac{(1+\widehat{y_b}^{(n)})}{2}\big) \bigg)/N.\!\!\! &&\!\!\triangleright\text{errors from }\fb  \label{eqn:error}
\end{align}

Next, we measure the interpretability of $\mR$, denoted as $\Omega(\mathcal{R})$. There exists a list of choices from the literature \citep{wang2017bayesian,lakkaraju2016interpretable, wang2018multi}. In this paper, we use the total number of conditions in rules, denoted as $\Omega(\mathcal{R})$. Regularizing the number of conditions not only increases interpretability but also naturally avoids overfitting. Without this regularization, the model will find long rules (having many conditions in a rule) with very small support but highly accurate, causing overfitting.  Therefore, regularizing the number of conditions will benefit the model in both model interpretability and predictive performance.  
The design of interpretability is customizable. Other metrics such as the overlap of rules, lengths of rules, etc., can also be used here.

Finally, we compute the transparency of a model. Following Definition \ref{def:exp}, the transparency of a hybrid rule set $\langle \mathcal{R}, \fb \rangle$ is 
\begin{equation}\label{eqn:exp1}
    \mathcal{E}(\langle\mathcal{R},\fb\rangle,\mathcal{D}) = \frac{\text{support}(\mathcal{R})}{|\mathcal{D}|}.
\end{equation}



We will write $\ell(\langle\mathcal{R},\fb\rangle,\mathcal{D})$ and $\mathcal{E}(\langle\mathcal{R},\fb\rangle,\mathcal{D}) $ as $\ell(\mathcal{R})$ and $\mathcal{E}(\mathcal{R}) $, respectively, ignoring the dependence on $\mathcal{D}$ and $\fb$ for simpler notations since $\fb$ and $\mathcal{D}$ are fixed.

Thus, we derive an objective function combining predictive accuracy, model interpretability and transparency
\begin{equation}
\Lambda(\mathcal{R}) = \ell(\mathcal{R}) + \alpha_1\Omega(\mathcal{R})-\alpha_2 \mathcal{E}(\mathcal{R}),
\end{equation}
 and our goal is to find an optimal model $\mathcal{R}^*$ such that
\begin{equation}\label{eqn:obj_rule}
\mathcal{R}^*= (\mR^*_+, \mR^*_-) \in \arg\min_{\mathcal{R}}    \Lambda( \mathcal{R}).
\end{equation}
Here, $\alpha_1, \alpha_2$ are non-negative coefficients. Tuning the parameters will produce models with different accuracy, interpretability, and transparency. For example, in an extreme case when $\alpha_2 >> \alpha_1$, the output will be a model that sends all data to $\fin$, producing a pure interpretable model. Then increasing $\alpha_1$ will produce a simpler and sparser model, using fewer rules and fewer conditions. When $\alpha_1>>\alpha_2$, the model will force $\fin$ to have no conditions and no rules, i.e., producing a pure black-box model. 

\subsection{Training with Adaptive Stochastic Local Search}
We describe a training algorithm to find an optimal solution $\mR^*$ to objective (\ref{eqn:obj_rule}). Learning rule-based models is challenging because the solution space is a power set of the rule space. Fortunately, our objective function suggests theoretical bounds that can be exploited for reducing computation. We first present the algorithm structure, then detail the theoretically grounded strategies we embed into the algorithm, and finally describe the proposing step.
\paragraph{Algorithm Structure} Given training examples, $\mathcal{D}$, black-box predictions $\{\widehat{y}_b^{(n)}\}_{n=1}^N$, parameters $\alpha_1,\alpha_2$,  base temperature $C_0$ and the total number of iterations $T$, we design an adaptive stochastic local search algorithm. Each state corresponds to a tuple of a positive rule set and a negative rule set, indexed by the time stamp $t$, denoted as $\mR^{[t]} = (\mR^{[t]}_+, \mR^{[t]}_-)$.  The starting state $\mR^{[0]}$ is initialized with two empty sets for $\mR_+^{[0]}$ and $\mR_-^{[0]}$, respectively. The temperature is a function of time $t$, $C_0^{1-\frac{t}{T}}$ and it decreases with time.  The neighboring states are defined as rule sets that are obtained via a small change to the current model. At each iteration, the algorithm improves one of the three terms (accuracy, interpretability, and transparency) with approximately equal probabilities, by removing a rule from the current model or adding a rule to the current model.
The proposed neighbor is then accepted with probability $\exp(\frac{\Lambda( \mathcal{R}^{[t]}) - \Lambda( \mathcal{R}^{[t+1]}) }{C_0^{1-\frac{t}{T}}})$ which gradually decreases as the temperature cools down till eventually converging to the final output. 


Next, we introduce the strategies and theoretical bounds to prune the search space before and during the search.
\paragraph{Rule Space Pruning} We first derive a bound to prune the search space before the algorithm begins. Assume the rules in $\mR^*$ are drawn from candidate rules. We use FP-growth\footnote{FP-Growth is an off-the-shelf rule miner. Other rule miners such as Apriori or Eclat can also be used.}  to generate two sets of candidate rules, $\Upsilon^+$ for positive rules and $\Upsilon^-$ for negative rules. The algorithm will search within this rule space so the complexity of the algorithm is directly determined by the size of the rule space. To reduce the search space, we derive a lower bound on the support of rules in $\mR^*$.
\begin{theorem}[Bound on Support]\label{theorem:support}
$\forall r \in \mR^*_+$, $\text{support}(r) \geq N\alpha_1$; 
$\forall r \in \mR^*_-, \text{support}(r) \geq\frac{N\alpha_1}{1-\alpha_2}$.
\end{theorem}
This means $\mR^*$ does not contain rules with too small a support. Rules that fail to satisfy the bounds can be safely deleted from the rule space without hurting the optimality. This theorem is used before the search begins to prune the rule space, which greatly reduces computation. On the other hand, removing rules with low support naturally helps prevent overfitting. The bound increases as $\alpha_1$ increases, since $\alpha_1$ represents the penalty for adding a rule. When $\alpha_1$ is large, the model is encouraged to use fewer rules, thus each rule need to cover enough instances in order to qualify for a spot in the model, therefore they are encouraged to have large support. All proofs are in the supplementary material.

\noindent\textbf{Search Chain Bounding} We also derive two bounds that dynamically prune the search space during the search. The bounds are applied in each iteration to confine the Markov Chain within promising solution space.
We first derive a bound on $\Omega(\mR^*)$, the total number of conditions in $\mR^*$.
Let $\lambda^{[t]}$ represent the best objective value found till time $t$, i.e.
$$
\lambda^{[t]} = \min_{\tau \leq t}\Lambda(\mR_{[\tau]}).
$$

We claim
\begin{theorem}[Bound on Interpretability]\label{theorem:size}
$\Omega(\mathcal{R}^*) \leq \frac{\lambda^{[t]} + \alpha_2}{\alpha_1}$.    
\end{theorem}
This theorem says that the number of conditions in $\mR^*$ is upper bounded. The theorem implies that the Markov Chain only needs to focus on the solution space of small models. Therefore, in the proposing step, if the current state violates the bound, neighbors with more conditions should not be proposed, equivalent to pruning the neighbors.

Next we derive an upper bound on the transparency with the similar purpose of confining the Markov Chain.
\begin{theorem}[Bound on Transparency]\label{theorem:transparency}
$ \text{support}(\mR^*) \geq \frac{N(\alpha_1 - \lambda^{[t]})}{\alpha_2}$.    
\end{theorem}
The theorem says the transparency of $\mR^*$ is lower bounded. Therefore when proposing the next state, if this theorem is violated, models with smaller transparency should not be proposed. 

Both bounds in Theorem \ref{theorem:size} and Theorem \ref{theorem:transparency} become smaller as $\lambda^{[t]}$ continuously gets smaller along the search steps. Exploiting the theorems in the search algorithm, the algorithm checks the intermediate solution at each iteration and pull the search chain back to the promising area (models of sizes smaller than $\frac{\lambda^{[t]} + \alpha_2}{\alpha_1}$ and transparency larger than $\frac{N(\alpha_1 - \lambda^{[t]})}{\alpha_2}$) whenever the bounds are violated, equivalent to pruning the search space at each iteration.

The algorithm is presented in Algorithm 1. Here we detail the proposing step.
 To propose a neighbor, at each iteration, we choose to improve one of the three terms (accuracy, interpretability, and transparency) with approximately equal probabilities. With probability $\frac{1}{3}$, or when the bound in Theorem~\ref{theorem:size} is violated, we aim to decrease the size of $\mathcal{R}^{[t]}$ (improve interpretability) by removing a rule from $\mR^{[t]}$ (line 8 - 9). With probability $\frac{1}{3}$ or when the lower bound on transparency (Theorem \ref{theorem:transparency}) is violated, we aim to increase the coverage of $\mR^{[t]}$ (improve transparency) by adding a rule to $\mR^{[t]}$ (line 10-11). Finally, with another probability $\frac{1}{3}$, we aim to decrease the classification error (improve accuracy) (line 13-26). 
To decrease the misclassification error, at each iteration, we draw an example from examples misclassified by the current model (line 13).  If the example is covered by $\mR^{[t]}$ (line 14), it means it was sent to the interpretable model but was covered by the wrong rule set.  
If the instance is negative (line 14), we remove a rule from the positive rule set that covers it. If the instance is positive, it means it is covered by the negative rule set. Then we either add a rule to the positive set to cover it or remove a rule in the negative rule set that covers it, re-routing it to $\fb$. If the example is not covered by $\mR^{[t]}$ (line 20), it means it was previously sent to the black-box model but misclassified, since we cannot alter the black-box model, we add a rule to the positive or negative rule set (consistent with the label of the instance) to cover the example.
When choosing a rule to add or remove, we first evaluate the rules using precision, which is the percentage of correctly classified examples in the rule. Then we balance between exploitation, choosing the best rule, and exploration, selecting a random rule, to avoid getting stuck in a local minimum.

\section{Hybrid \emph{Linear Models}}\label{sec:linear}
Next, we instantiate the hybrid framework with linear models and call the model Hybrid Linear Model. Our goal is to learn $\fin=f_\text{linear}=(\mathbf{w},\theta_+,\theta_-)$, where $\mathbf{w}$ is the coefficient of a linear model and $\theta_+$ and $\theta_-$ ( $\theta_+\geq \theta_-$) are two thresholds, in order to form $f = \langle f_\text{linear}, \fb \rangle$  as:
\begin{eqnarray}
\label{eq:hybridf}
f(\mathbf{x})=
\left\{
\begin{array}{cl}
     1&  \text{ if } \mathbf{w}^\top\mathbf{x} \geq \theta_+\\
     -1& \text{ if } \mathbf{w}^\top\mathbf{x} \leq \theta_-\\
     f_b(\mathbf{x})& \text { otherwise}
\end{array}
\right.
\end{eqnarray}
Similar to a hybrid rule set, a hybrid linear model partitions the data space into three regions, ``certainly positive'', ``certainly negative'' and an undetermined region left to the black-box model. 
\vspace{-2mm}
\subsection{Model Formulation}
We cast the optimization for a hybrid linear model into the hybrid decision-making framework and re-formulate the learning objective.

\subsubsection*{Learning Objective}
The (in-sample) predictive performance characterizes the fitness of the model to the training data. Since $\fb$ is pre-given, the predictive performance is determined by two factors, the accuracy of $f_\text{linear}$ on instances with $\mathbf{w}^\top\mathbf{x} \geq \theta_+$ or $\mathbf{w}^\top\mathbf{x} \leq \theta_-$ and the accuracy of $\fb$ on the remaining examples. We wish to obtain a good partition of data $\mathcal{D}$ by assigning $\fb$ and $f_\text{linear}$ to different region of the data such that the strength of $\fb$ and $f_\text{linear}$ are properly exploited. Second, we include the gap $\theta_+ - \theta_-$ as a penalty term in the objective to account for transparency of a hybrid linear model. The smaller gap between $\theta_+$ and $\theta_-$ there is, more data are classified by the linear model. In the most extreme case where $\theta_+ - \theta_- = 0$, all data are sent to the linear model, and the hybrid linear model is reduced to a pure linear classifier, i.e., transparency equals one. 
Finally, we also need to consider model regularization in the objective. As the weight for the sparsity enforcing regularization term increases, the model encourages using a smaller number of features which increases the interpretability of the model as well as preventing overfitting.

Combining the three factors discussed above, we formulate the learning objective for hybrid linear model as: 
\begin{equation}
\label{eq:obj}
F^*:=\min_{\mathbf{w},\theta_+\geq \theta_-}\left\{F(\mathbf{w},\theta_+,\theta_-):=\mathcal{L}(\mathbf{w},\theta_+,\theta_-;\fb)  + \alpha_1 r(\mathbf{w})+ \alpha_2 (\theta_+ - \theta_-)\right\}, 
\end{equation}
where $\mathcal{L}(\mathbf{w},\theta_+,\theta_-;\fb)$ is the loss function defined on the training set, $\theta_+ - \theta_-$ is a penalty term to increase the transparency of $f$, $r$ is a convex and closed regularization term (e.g. $\|\mathbf{w}\|_1$, $\frac{1}{2}\|\mathbf{w}\|_2^2$ or an indicator function of a constraint set), and $\alpha_1$ and $\alpha_2$  are non-negative coefficients which balance the importance of the three components in~\eqref{eq:obj}. 

\begin{figure}[h]
\centering
  \includegraphics[width=0.45\linewidth]{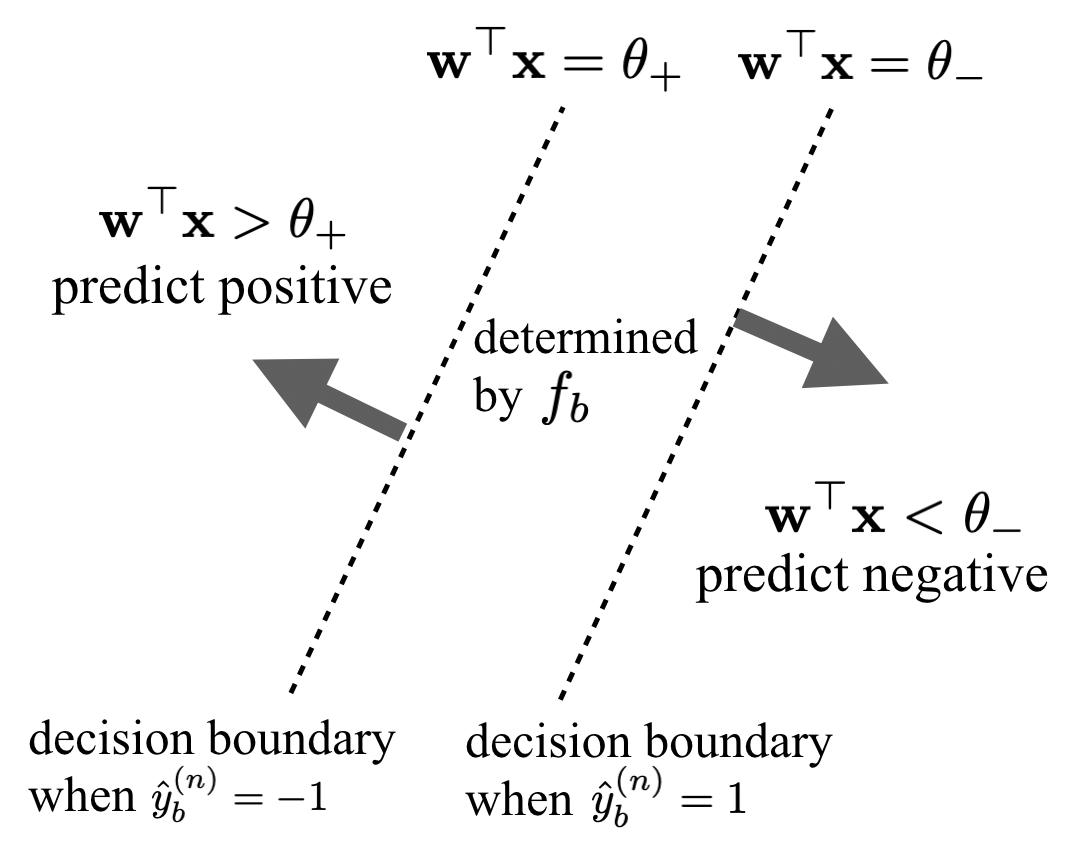}
\caption{The partitioning of the data space by two decision boundaries}\label{fig:dataspace}
\end{figure}

Next, we define $\mathcal{L}(\mathbf{w},\theta_+,\theta_-;\fb)$ in details. According to \eqref{eq:hybridf}, the hybrid model $f$ will misclassify an instance $\mathbf{x}^{(n)}$ in two scenarios. When $y_b^{(n)} = 1$,  $\mathbf{w}^\top\mathbf{x} = \theta_-$ becomes the active decision boundary in the sense that $\mathbf{x}^{(n)}$ with $\mathbf{w}^\top\mathbf{x}^{(n)} \geq \theta_-$ is predicted as positive (if $\mathbf{w}^\top\mathbf{x}^{(n)} \geq \theta_+$, it is predicted as positive by the linear model; if $\theta_-\leq\mathbf{w}^\top\mathbf{x}^{(n)} \leq \theta_+$, it is predicted as positive by the black-box since $y_b^{(n)} = 1$) and $\mathbf{x}^{(n)}$ with  $\mathbf{w}^\top\mathbf{x} < \theta_-$ is classified as negative according to \eqref{eq:hybridf}. When $y_b^{(n)} = -1$, $\mathbf{w}^\top\mathbf{x} = \theta_+$ becomes the active decision boundary in the sense that $\mathbf{x}^{(n)}$ with $\mathbf{w}^\top\mathbf{x}^{(n)} \geq \theta_+$ is predicted as positive and $\mathbf{x}^{(n)}$ with  $\mathbf{w}^\top\mathbf{x}^{(n)} < \theta_+$ is classified as negative according to \eqref{eq:hybridf} (if $\mathbf{w}^\top\mathbf{x}^{(n)} \leq \theta_-$, it is predicted as negative by the linear model; if $\theta_-\leq\mathbf{w}^\top\mathbf{x}^{(n)} \leq \theta_+$, it is predicted as negative by the black-box since $y_b^{(n)} = -1$). See Figure~\ref{fig:dataspace} for an illustration of the decision boundaries. 



To define the losses associated to $f$ in these two scenarios, we partition the data into two subsets based on the labels predicted by the black-box $\{y_b^{(n)}\}_{n=1}^N$. In particular, we define
\begin{equation}
I^+_b = \{n|y_b^{(n)}=1\}\text{ and } I^-_b = \{n|y_b^{(n)}=-1\}.
\end{equation}
The loss function in \eqref{eq:obj} over the dataset $\mathcal{D}$ is then defined as 
\begin{equation}
\label{eq:loss}
\mathcal{L}(\mathbf{w},\theta_+,\theta_-;\fb) =
\frac{1}{N} \sum_{n\in I^+_b} \phi(y^{(n)}(\mathbf{w}^\top \mathbf{x}^{(n)} - \theta_-)) +\frac{1}{N}\sum_{n\in I^-_b} \phi(y^{(n)}(\mathbf{w}^\top \mathbf{x}^{(n)} - \theta_+)),
\end{equation}
where $\phi(z):\mathbb{R}\rightarrow \mathbb{R}$ is a non-increasing convex closed loss function which can be one of those commonly used in linear classification such as 
\begin{equation}
\label{eq:lossphi}
\left\{
    \begin{array}{rl}
       \text{hinge loss}  & \phi(z)=(1-z)_+ \\
    \text{smoothed hinge loss} & \phi(z)=\frac{1}{2}(1-z)_+^2\\
    \text{logistic loss} & \phi(z)=\log(1+\exp(-z)).
    \end{array}
    \right.
\end{equation}
Note that $I^+_b$ and $I^-_b$ form a partition of $\{1,2,\dots,N\}$ and each data point corresponds to one loss term in \eqref{eq:loss}. The intuition of this loss function is as follows. Take a data point $\mathbf{x}^{(n)}$ with $y^{(n)} = 1$ and $y_b^{(n)} = 1$ as an example. Our hybrid model \eqref{eq:hybridf} will classify $\mathbf{x}^{(n)}$ correctly as long as $\mathbf{x}^{(n)}$ does not fall in the ``negative half space'', namely, $\mathbf{w}^\top \mathbf{x}^{(n)} > \theta_-$. Hence, with the non-increasing property of $\phi$, the loss term $\phi(y^{(n)}(\mathbf{w}^\top \mathbf{x}^{(n)} - \theta_-))$ will encourage a positive value of $\mathbf{w}^\top \mathbf{x}^{(n)} - \theta_-$. On the other hand, for a data point $\mathbf{x}^{(n)}$ with $y^{(n)} = 1$ and $y_b^{(n)} = -1$, our hybrid model will classify $\mathbf{x}^{(n)}$ correctly only when $\mathbf{x}^{(n)}$ falls in the ``positive half space'', namely, $\mathbf{w}^\top \mathbf{x}^{(n)} \geq \theta_+$. We use the loss term $\phi(y^{(n)}(\mathbf{w}^\top \mathbf{x}^{(n)} - \theta_+))$ to encourage a positive value of $\mathbf{w}^\top \mathbf{x}^{(n)} - \theta_+$. The true label $y^{(n)}=1$ in both examples above but the similar interpretation applies to the case of $y^{(n)}=-1$.

\vspace{-3mm}
\subsection{Model Training}
With the loss function defined in \eqref{eq:loss}, the hybrid model can be trained by solving the convex minimization problem~\eqref{eq:obj} for which many efficient optimization techniques are available in literature including subgradient methods~\citep{nemirovski2009robust,duchi2011adaptive}, accelerated gradient methods~\citep{nesterov2013introductory,beck2009fast}, primal-dual methods~\citep{nemirovski2004prox,chambolle2011first} and many stochastic first-order methods based on randomly sampling over coordinates or data~\citep{johnson2013accelerating,shalev2014accelerated,duchi2011adaptive,zhang2017stochastic}. 
The choice of algorithms for~\eqref{eq:obj} depends on various characteristics of the problem such as smoothness, strong convexity and data size. 
\begin{algorithm}[t]
\label{alg:APG}
\caption{Smoothing Accelerated Proximal Gradient Method for \eqref{eq:obj}  \citep{nesterov2007smoothing,nesterov2013introductory}}\label{alg:APG}
\small
\begin{algorithmic}[1]
\STATE \textbf{Input}: The initial solution $\mathbf{w}^{[0]}\in \text{dom}(r)$, $(\theta_+^{[0]},\theta_-^{[0]})$ satisfying $\theta_+^{[0]}\geq \theta_-^{[0]}$, step length parameters $\eta>0$ and $\alpha_0\in(0,1)$, the smoothing parameter $\mu\geq0$ and the number of iterations $T$.
\STATE Set $(\widehat{\mathbf{w}}^{[0]},\widehat\theta_+^{[0]},\widehat\theta_-^{[0]})=({\mathbf{w}}^{[0]},\theta_+^{[0]},\theta_-^{[0]})$
\FOR{$t=1,2,\ldots,T$}
\STATE 
\begin{eqnarray*}
\mathbf{w}^{[t]}&=&\argmin\limits_{\mathbf{w}\in\mathbb{R}^d} \left\langle \nabla_{\mathbf{w}}\mathcal{L}_{\mu}(\widehat{\mathbf{w}}^{[t-1]},\widehat\theta_+^{[t-1]},\widehat\theta_-^{[t-1]}),
\mathbf{w}\right\rangle \notag +\frac{1}{2\eta}\left\|\mathbf{w}-\widehat{\mathbf{w}}^{[t-1]}\right\|_2^2 +\alpha_1r(\mathbf{w})\\
(\theta_+^{[t]},\theta_-^{[t]})&
=&\argmin\limits_{\theta_+\geq\theta_-}\left\{
\begin{array}{c}
[\nabla_{\mathbf{\theta_+}}\mathcal{L}(\widehat{\mathbf{w}}^{[t-1]},\widehat\theta_+^{[t-1]},\widehat\theta_-^{[t-1]})+\alpha_2]\theta_+\\
+[\nabla_{\mathbf{\theta_-}}\mathcal{L}(\widehat{\mathbf{w}}^{[t-1]},\widehat\theta_+^{[t-1]},\widehat\theta_-^{[t-1]})-\alpha_2]\theta_-\\
+\frac{1}{2\eta}\left\|\left(\begin{array}{c}\theta_+\\\theta_-\end{array}\right)-\left(\begin{array}{c}\widehat\theta_+^{[t-1]}\\\widehat\theta_-^{[t-1]}\end{array}\right)\right\|_2^2 \end{array}\right\}
\end{eqnarray*}
\STATE Compute $\alpha_{t}\in(0,1)$ from the equation $\alpha_{t}^2=(1-\alpha_{t})\alpha_{t-1}^2$
and set $\beta_{t}=\frac{\alpha_{t-1}(1-\alpha_{t-1})}{\alpha_{t-1}^2+\alpha_{t}}$.
\STATE 
$$
\left(\begin{array}{c}\widehat{\mathbf{w}}^{[t]}\\\widehat\theta_+^{[t]}\\\widehat\theta_-^{[t]}\end{array}\right)=
\left(\begin{array}{c}{\mathbf{w}}^{[t]}\\\theta_+^{[t]}\\\theta_-^{[t]}\end{array}\right)+\beta_t\left[\left(\begin{array}{c}{\mathbf{w}}^{[t]}\\\theta_+^{[t]}\\\theta_-^{[t]}\end{array}\right)-\left(\begin{array}{c}{\mathbf{w}}^{[t-1]}\\\theta_+^{[t-1]}\\\theta_-^{[t-1]}\end{array}\right)\right]
$$
\ENDFOR
\STATE  \textbf{Output: } $\mathbf{w}^{[t]},\theta_+^{[t]},\theta_-^{[t]}$
\end{algorithmic}
\end{algorithm}
\normalsize

Since numerical optimization is not the focus of this paper, we will simply utilize the accelerated proximal gradient method (APG) by \cite{nesterov2013introductory} to solve \eqref{eq:obj} when $\phi$ is smooth (e.g. the second and third cases in \eqref{eq:lossphi}). 
When $\phi$ is non-smooth (e.g. the first cases in \eqref{eq:lossphi}), we apply the well-known smoothing technique~\citep{nesterov2007smoothing} to create a smooth approximation for $\phi$ and then solve the smooth problem using APG. To describe the smoothing technique, we denote by $\phi^*(\alpha):\mathbb{R}\rightarrow \mathbb{R}\cup\{+\infty\}$ the Fenchel conjugate of $\phi$, namely,  $\phi^*(\alpha)=\sup_{z}\{\alpha z- \phi(z)\}$. It is well-known in convex analysis that, given our assumptions on $\phi$, we have $\phi(z)=\sup_{\alpha}\{z\alpha - \phi^*(\alpha)\}$.
Let $h(\alpha):\mathbb{R}\rightarrow \mathbb{R}$ be a $1$-strongly convex function with respect to the Euclidean norm. Given a smoothing parameter $\mu\geq0$, the smooth approximation for $\phi$ can be constructed as
$\phi_{\mu}:=\sup_{\alpha}\{z\alpha - \phi^*(\alpha)-\mu h(\alpha)\}$. According to \cite{nesterov2007smoothing}, the function $\phi_{\mu}$ is smooth and its gradient is $\nabla\phi_{\mu}(z)=\arg\min_{\alpha}\{z\alpha - \phi^*(\alpha)-\mu h(\alpha)\}$ which has a closed-form for most commonly used $\phi$. According to \cite{nesterov2007smoothing}, we have the relationship $\phi_{\mu}(z)\leq\phi(z)\leq \phi_{\mu}(z)+\mu\max_{\alpha\in\text{dom}(\phi^*)}h(\alpha)$ where $\text{dom}(\phi^*)$ is typically compact for non-smooth $\phi$.\footnote{In this paper, we use $\text{dom}(f)$ to represent the domain of a function $f$.} These facts suggest that, to solve \eqref{eq:obj} with non-smooth $\phi$, we can approximate $\phi$ with $\phi_{\mu}$ and then apply the APG method \citep{nesterov2013introductory} to the smooth problem
\begin{equation}
\label{eq:sobj}
\min_{\mathbf{w},\theta_+\geq \theta_-} \mathcal{L}_{\mu}(\mathbf{w},\theta_+,\theta_-;\fb)+ \alpha_1 r(\mathbf{w}) + \alpha_2 (\theta_+ - \theta_-) ,
\end{equation}
where \vspace{-2mm}
\begin{equation}
\label{eq:sloss}
\mathcal{L}_{\mu}(\mathbf{w},\theta_+,\theta_-;\fb) =
\frac{1}{N} \sum_{n\in I^+_b} \phi_{\mu}(y^{(n)}(\mathbf{w}^\top \mathbf{x}^{(n)} - \theta_-))\notag+\frac{1}{N}\sum_{n\in I^-_b}\phi_{\mu}(y^{(n)}(\mathbf{w}^\top \mathbf{x}^{(n)} - \theta_+)).
\end{equation}
The returned solution for  \eqref{eq:sobj} will also be a good solution to \eqref{eq:obj} when $\mu$ is small. Note that when $\mu=0$, we have $\phi_{\mu}(z)=\phi(z)$ and \eqref{eq:sobj} and \eqref{eq:sloss} are reduced to \eqref{eq:obj} and \eqref{eq:loss}. This smoothing APG method is presented as Algorithm \ref{alg:APG}. When $\phi$ is smooth, one just needs to choose $\mu=0$. For simplicity of notation, we have suppressed the dependence of $\mathcal{L}_{\mu}(\mathbf{w},\theta_+,\theta_-;\fb)$ on $\fb$.

The two subproblems in Step 4 and Step 5  in Algorithm \ref{alg:APG} can be solved in closed forms for most commonly used $r$ (e.g., $r(\mathbf{w})=\|\mathbf{w}\|_1$). In Algorithm \ref{alg:APG},  users need to provide a step length parameter $\eta$. In order to guarantee the convergence  in theory, this parameter can be chosen as $\eta=\frac{1}{L_{\mu}}$. Here, $L_{\mu}$ is the Lipchitz continuity constant\footnote{The explicit expression of $L_{\mu}$ can be found in ~\cite{nesterov2007smoothing}.} of $\nabla\mathcal{L}_{\mu}$ which is in the order of $O(\frac{1}{\mu})$ if $\phi$ is non-smooth and $\mu>0$ and is identical to the Lipchitz continuity constant $\nabla\mathcal{L}$ if $\phi$ is smooth and $\mu=0$. When $\eta$ is chosen to be this theoretical value, Algorithm \ref{alg:APG} can ensure $F(\mathbf{w}^{[t]},\theta_+^{[t]},\theta_-^{[t]})-F^*\leq \epsilon$ in $T=O(\frac{1}{\epsilon})$ iterations when $\phi$ is non-smooth and in $T=O(\frac{1}{\sqrt{\epsilon}})$ iterations when $\phi$ is smooth. See~\cite{nesterov2007smoothing,nesterov2013introductory} for the details of the convergence analysis. Although a theoretical value for $\eta$ has a closed form, it is generally too conservative and leads to slow convergence in practice. To address this issue, a line search scheme, for example the one proposed in \cite{lin2015adaptive}, can be applied to adaptively choose $\eta$ during the iterations, which can improve the convergence speed in practice significantly while still preserve the theoretical convergence property. We implement APG with this adaptive line search scheme in our numerical experiments.

\section{Experiments}\label{sec:experiments}
We perform a detailed experimental evaluation of hybrid models on public data from domains where interpretability is critical, including healthcare, judiciaries, and business problems. On these datasets, the two forms of interpretable models will collaborate with popular black-box models including ensemble models and neural networks and solve problems on  structured data and text data. We first discuss the evaluation metrics, then describe experiments setup and finally analyze the experimental results.

\subsection{Evaluation Metrics}
The goal for building a hybrid model is to provide interpretable and accurate predictions on as many data points as possible. Thus we propose  two metrics below to evaluate the predictive performance, interpretability, and transparency.

\noindent\textbf{Model Complexity} The two types of models we choose (rules and linear models) have naturally interpretable forms. 
To provide a consistent evaluation of model complexity for hybrid rule sets and hybrid linear models, we measure the number of smallest \textbf{cognitive units} in a model. For rule sets, we count the total number of conditions (feature-value pairs) in association rules. Note that we choose the total number of conditions instead of the number of features because a feature could be used multiple times by different rules in a model. For linear models, we count the number of non-zero coefficients. Both of them represent the amount of units a user needs to comprehend in order to understand a model. 

\begin{figure}[h!]
\centering
  \includegraphics[width=0.5\textwidth]{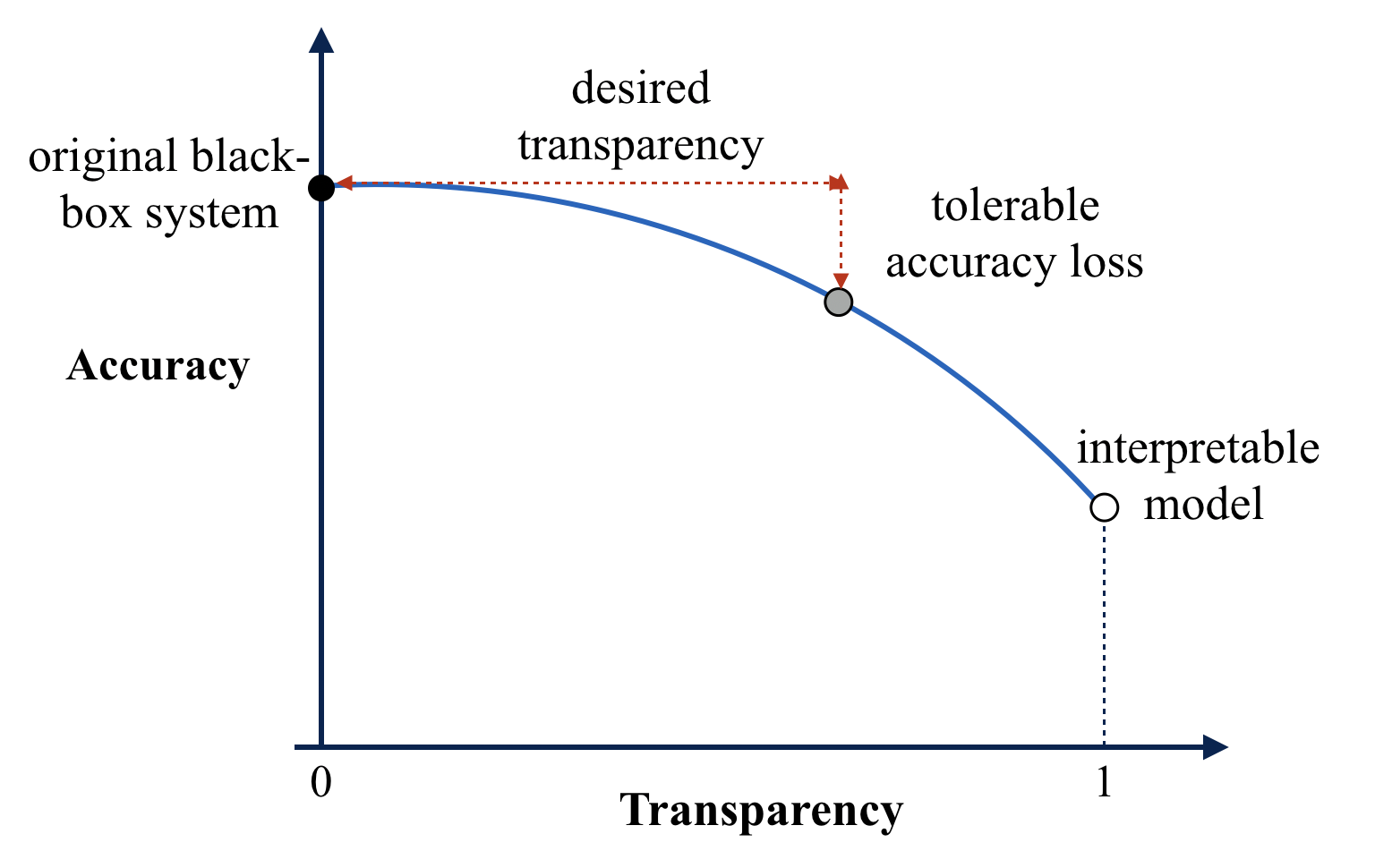}
\caption{Efficient frontier of a hybrid model.}\label{fig:efficient}    
\end{figure}

\noindent\textbf{Efficient Frontier}  To evaluate the trade-off between transparency and accuracy, we propose efficient frontiers. An Efficient Frontier represents how much accuracy a model needs to trade to obtain a certain level of transparency. See Figure \ref{fig:efficient} for an illustration. The curves provide a practical and straightforward way for model selection. Instead of choosing between two discrete choices of either black-box models or interpretable models, now users can find a middle ground where partial transparency and good predictive performance can co-exist, providing one with a wider range of models. Users can choose the best operating point on this curve, based on one's desired transparency as well as the tolerable loss in predictive performance.
\subsection{Structured Datasets}
\label{sec:struct}
We use four structured datasets that are publicly available. 1) \emph{census} \citep{kohavi1996scaling} (48,842 $\times$ 14)  predicts if an indivisual's annual income is greater than $50K$ based on the individual's demographic information, education background, and work information. 2) \emph{juvenile}\citep{osofsky1995effect} (4,023 observations and 55 reduced features) studies the consequences of juvenile exposure to violence. The dataset was collected via a survey. 3) \emph{coupon} (2017 $\times$ 90 binarized features) predicts if a customer will accept a coupon recommended by an in-vehicle recommender system. 4) \emph{stop-and-frisk} (80,755 $\times$ 27) predicts if someone will be frisked by a police officer, using data from NYPD Stop, Question, and Frisk database from 2014 to 2016.

\paragraph{Baselines}
 We process each dataset by binarizing all categorical features. For each structured dataset, we build three canonical black-box models as baselines and also later as input for hybrid models, Random Forests \citep{liaw2002classification}, AdaBoost \citep{freund1995desicion} and extreme gradient boosting trees (XGBoost) \citep{chen2016xgboost}. 
 We partition each dataset into 80\% training and 20\% testing. We do 5-fold cross-validation for parameter tuning on the training set to choose the best parameters and then evaluate the best model on the test set. To provide a baseline for interpretability and accuracy comparison, we also build interpretable models. The methods we choose here include traditional decision trees CART \citep{breiman2017classification}, C4.5 \citep{quinlan2014c4} and C5.0 \citep{pandya2015c5} and more recent rule-based classifiers, Bayesian Rule Sets (BRS) \citep{wang2017bayesian} and Scalable Bayesian Rule List \citep{yang2017scalable}. BRS and SBRL are two recent representative methods which have proved to achieve simpler models with competitive predictive accuracy compared to the older rule-based classifiers.  We also include Lasso \citep{tibshirani1996regression} to provide a baseline for hybrid linear models. 
 
 \paragraph{Implementations}
We use R  packages \citep{hornik2007rweka,quinlan2004data} to build interpretable methods except for BRS which has the code publicly available\footnote{https://github.com/wangtongada/BOA}. For C4.5 and C5.0, we tune the minimum number of samples at splits. For BRS, we set the maximum length of rules to 3 as recommended by the paper \citep{wang2017bayesian}. For BRS, there are parameters $\alpha_+,\beta_+,\alpha_-,\beta_-$ that govern the likelihood of the data. We set $\beta_+, \beta_-$ to 1 and vary $\alpha_+,\alpha_-$ from $\{100,1000,10000\}$.  For SBRL, there are hyperparameters $\lambda$ for the expected length of the rule list and $\eta$ for the expected cardinality of the rules in the optimal rule list.  We vary $\lambda$ from $\{5,10,15,20\}$ and $\eta$ from $\{1,2,3,4,5\}$. To preserve interpretability, all interpretable baselines have the number of rules under 30 (for decision trees we count the number of leaves).  Then we use python packages \citep{pedregosa2011scikit} for the black-box models.  For random forest, we tune the number of trees and the number of features for each tree. For Adaboost, we set the number of estimators to 800 and tune the maximum number of features and the maximum depth of trees. For XGBoost, we set the number of base learners to be 800 and set the learning rate to 0.01. Then we tune the maximum depth of trees, maximum features used by a tree and the minimum samples that need to exist in a leaf. For Lasso, we tuned the regularizing parameter $\lambda$. Models with the best cross-validated validation performance are selected.
 
\begin{figure}[h!]
\centering
  \includegraphics[width=\textwidth]{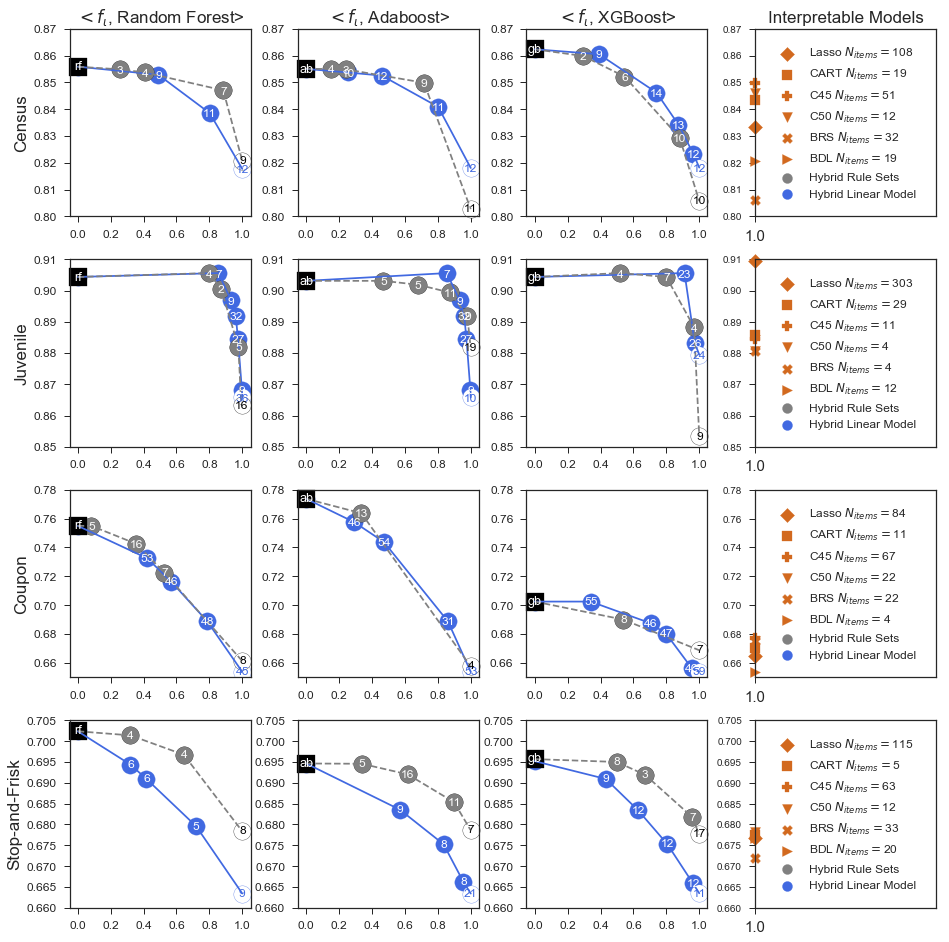}\vspace{-2mm}
\caption{Transparency efficient frontiers of hybrid models. The X-axis represents the transparency, and the Y-axis represents the accuracy on the test set. The value inside each circle is the number of items in a hybrid model. ``rf'', ``ab'' and ``gb'' in the black squares represent Random Forest, Adaboost, and XGBoost.}\label{fig:exp1}   \end{figure}

 The predictive performance of the black-box and interpretable models are reported in Table \ref{tab:models}. On all four datasets, black-box models outperform interpretable models with one exception, Lasso on juvenile dataset. But Lasso uses 303 non-zero coefficients in this model, making it hardly interpretable. To compare interpretability against the hybrid models, we report the number of conditions for rule-based models and the number of non-zero coefficients for Lasso in Figure \ref{fig:exp1}. 
   
 \begin{table}[h!]
 \small
 \caption{Accuracy of interpretable models on test sets }\label{tab:models}
\begin{tabular}{lccc|cccccc}
\toprule
\multirow{2}{*}{Datasets} & \multicolumn{3}{c|}{\emph{Black-box Models}} & \multicolumn{6}{c}{\emph{Interpretable Models}} \\  \cline{2-10} 
 & RF  & Adaboost  & XGBoost & CART   & C4.5   & C5.0   & BRS   & SBRL & Lasso  \\ \hline 
\textbf{Census} &0.856 &0.856 &0.862 & 0.843       & 0.850       &  0.846      & 0.806      &0.820    &  0.833  \\
\textbf{Juvenile} & 0.904 & 0.903 & 0.904  & 0.886& 0.884&0.881        & 0.881     & 0.886   &  0.909  \\
 \textbf{Coupon}& 0.755& 0.774& 0.703&   0.670      & 0.678  &0.675     & 0.675      &  0.654  & 0.665  \\
 \textbf{Stop\&Frisk}& 0.702 & 0.695  &0.696 &0.678 &0.677        &0.679 &  0.672     &   0.677  & 0.677\\ \bottomrule
\end{tabular}
\end{table}

   \paragraph{Building Hybrid Models}
   Given the black-box models built above, we create hybrid rule sets and hybrid linear models. 
 For hybrid rule sets, we set the maximum length of rules to be 4. We choose $\alpha_1$ from $[0.001,0.06]$ and choose $\alpha_2$ from $[0.001, 0.5]$ to obtain a list of models. For hybrid linear models, we choose $\alpha_1$ from $[0.02,0.06]$ in order to select a small number of features. We choose $\alpha_2$ from $[0.001, 0.5]$ to obtain a list of models. When building the hybrid linear model for all datasets, we choose $\phi$ to be the hinge loss in \eqref{eq:lossphi}, choose $r(x)=\|x\|_1$, and run Algorithm~\ref{alg:APG} for $5,000$ iterations with $\mu=0.0001$. The changes of the objective value in Algorithm~\ref{alg:APG} is less than 0.01\% in the last iterations, indicating the convergence of the algorithm.
   
   We report the run time in Table \ref{tab:runtime}. Hybrid linear models are much more efficient since the search algorithm is based on gradient descent and computationally convenient. Hybrid rule sets, on the other hand, involve evaluating intermediate solutions and proposing next states, and therefore they are computationally heavier.
   \begin{table}[h]
   \caption{Means and standard deviations of training time for hybrid models}\label{tab:runtime}
\begin{tabular}{l|llll}
\toprule
 & Census & Juvenile & Coupon & Stop-and-Frisk \\ \hline
Hybrid Rule Set &  10m 55s $\pm$ 45s &2m 10s $\pm$ 21s     &  3m 25s $\pm$ 6s        & 41m 50s $\pm$ 5s  \\
Hybrid Linear Model &   1m 5s $\pm$ 1s    &    26s $\pm$ 1s   &  52s $\pm$ 2s      &  2m 32s $\pm$ 3s \\ \bottomrule
\end{tabular}
\end{table}

\paragraph{Efficient Frontiers Analysis}
 We plot the transparency efficient frontiers for $\langle f_\text{rules}, \fb \rangle$ and $\langle f_\text{linear}, \fb \rangle$ in Figure \ref{fig:exp1}, evaluated on test sets. For easier comparison, we set the range of the Y-axis to be the same for each dataset (row).   The black-box baseline models are located at transparency equal to zero in each plot, represented by a black square. The hybrid models span the entire spectrum of transparency with accuracy slowly decreasing as transparency increases  till reaching an interpretable model at transparency equal to one, represented by an empty circle. Besides, we show baseline interpretable models at the same row for each dataset. All of them are located at transparency equal to one. 
 
 The efficient frontiers are concave with different decaying rates for the four datasets. For the census dataset, the curves remain almost flat as transparency is less than 50\%. This suggests that transparency is gained at nearly no cost of predictive performance if the transparency is not too large. In addition, this \textbf{free} transparency is obtained using very few conditions for hybrid rule sets, only 4 or 5 conditions (1 or 2 rules in total), compared to the interpretable baselines using a lot more conditions but achieving lower accuracy overall. For the juvenile dataset, it is interesting to observe the accuracy even increases, although slightly, when 80\% of the data are sent to an interpretable model, then drops rapidly if continuing sending data to interpretable models. This means the majority of the data can be easily explained by a couple of short rules or simple linear models, and the rest 20\% need to be handled by a more advanced black-box model. If forcing everything to interpretable models, not only the performance will decrease, but the model complexity will also increase significantly. We notice that for this dataset, the highest accuracy is achieved by Lasso, 0.91. But Lasso uses 303 non-zero coefficients, turning it into a practically black-box model for having so many cognitive units. The results on census and juvenile datasets proved our hypothesis that even a black-box model is globally better, there might exist a subspace where a very simple model can do just as well or even slightly better, since simpler models are less likely to overfit. Finally, for the last two datasets, coupon recommendation and stop-and-frisk prediction, the efficient frontier shows a clear trade-off between transparency and accuracy, with the accuracy steadily decreases as transparency increases.
 
We observe that hybrid linear models generally use more cognitive units (non-zero coefficients) in a model than hybrid rule sets (conditions).  This is because linear models need more features to accurately define a decision boundary in the feature space. A slight change in the coefficients may affect the partitioning of the entire data. Therefore, regularizing the number of coefficients too much  will hurt the performance. Thus the model has to keep many non-zero coefficients.
On the other hand, hybrid rule sets use association rules, which are hyper-cubes in the data space and each are defined by a couple of features only. It has been shown in the literature that allowing rules to have length up to 4 can already accurately place it on correct subsets and achieve high predictive performance \citep{wang2017bayesian}. 
 
Below we show examples of hybrid models from two datasets.
\subsubsection{Example 1: Hybrid Models for the Census Dataset}
The census dataset predicts if an individual's annual income is greater than \$50K, based on his/her demographic information (gender, age, marital status, race, etc), education background (degree and the number of years of education) and work information (occupation, hours for work per week, etc). 
\begin{table}[h]
\centering
\footnotesize
\caption{$\langle f_\text{rule}, \text{Adaboost}\rangle$ for juvenile delinquency dataset. The transparency is 71.6\% and the accuracy is 85.0\%, evaluated on the test set.}\label{tab:adult_rs}
\begin{tabular}{llc}
\toprule
        & \multicolumn{1}{c}{\textbf{Rules}}                                                             & \multicolumn{1}{c}{\textbf{Model}}             \\\hline
\textbf{If} &  capital gain $= 0$ \textbf{\emph{and}} capital loss $= 0$ \textbf{\emph{and}} number of years of education $\leq$ 13 & \\
& \textbf{\emph{and}} degree $\neq$ Bachelor's \textbf{\emph{and}} occupation $\neq$ Tech-support   &\multirow{2}{*}{$\Bigg\} \mathcal{R}_-$} \\
& \textbf{\emph{OR}} age $\leq$ 41 \textbf{\emph{and}} capital gain = 0 \textbf{\emph{and}} hours of work per week $<$ 40 & \\
&\textbf{\emph{and}} marital status $\neq$ Married-civ-spouse & \\
        & $\rightarrow$ \textcolor{red}{$Y = -1$} (income $<$ \$50K) & \\ 
\textbf{Else}    &\textcolor{blue}{$Y = \fb(\mathbf{x})$} &  \multicolumn{1}{c}{$f_b$}      \\ \bottomrule                
\end{tabular}\vspace{-2mm}
\end{table}
 \begin{figure}[h!]
\centering
  \includegraphics[width=0.6\textwidth]{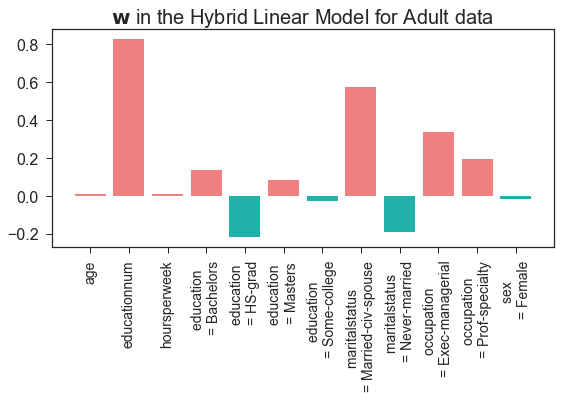}\vspace{-2mm}
\caption{Coefficients in a $\langle f_\text{linear}, \text{Adaboost} \rangle$ model for Census dataset. The model has transparency of 39.0\% and accuracy of 86.1\%. $\theta_+ = 1.98, \theta_- = 0.70$.}\label{fig:adult_linear}   
\end{figure}
We choose a hybrid rule set model that substitute Adaboost with reasonably large transparency and accuracy. We show the model in Table \ref{tab:adult_rs}. The positive rule set is empty in this model. The negative rule set consists of two rules that capture ``certainly negative'' instances. They cover 71.6\% of the data and the overall accuracy is 85.0\%. 

We also show a hybrid linear model that also substitutes Adaboost model in Figure \ref{fig:adult_linear}. The model identifies the most positive features include the number of years of education, if the marital status is married to a civilian spouse, and if the occupation is executive managerial or professional specialty; the most negative features are if education is high-school graduate and the marital status is never married.
\subsubsection{Example 2: Hybrid Models for the Juvenile Deliquency Dataset}
We show hybrid models built from the juvenile delinquency data. The dataset was collected from a survey where children were asked questions regarding their exposure to violence from family members, friends, or other sources in their community and use that the predict if s/he will commit delinquency. Therefore, each feature is a question from the survey and the corresponding values are respondents' answer to the questions. 
For example, question 18 asks ``has anyone (including family members or friends) ever attacked you with a gun, knife or some other weapon?'' and the answer is ``Yes'', ``No'', ``Refused to answer'', or ``Not sure''.

We first show rule sets substitutes in an XGBoost model in Table \ref{tab:juvenile_rs}. The positive rule set contains one rule with three conditions, and the negative rule set contains one rule with four conditions. They collectively cover and explain 80\% of the data without losing any predictive performance from XGBoost alone.
\begin{table}[h]
\footnotesize
\caption{$\langle f_\text{rule}, \text{XGboost}\rangle$ for juvenile delinquency dataset. The transparency is 80.0\% and the accuracy is 90.6\%, evaluated on the test set.}
\label{tab:juvenile_rs}
\begin{tabular}{llc}
\toprule
        & \multicolumn{1}{c}{\textbf{Rules}}                                                             & \multicolumn{1}{c}{\textbf{Model}}             \\\hline
\textbf{If} &  has anyone (including family members or friends) ever attacked you with a gun,  &\multirow{5}{*}{$\Bigg\} \mathcal{R}_+$} \\
&  knife or some other weapon? (Q18A) $=$ ``Yes'' \textbf{\emph{and}} (Q18C) has anyone (including  & \\
& family members or friends) ever threatened you with  a gun  or knife, but didn't  & \\
& actually shoot or cut you $=$ ``Yes'' \textbf{\emph{and}}  has your friends ever purposely damaged & \\
& or destroyed property that did not belong to them ? (Q48AA)$=$ ``Yes'' & \\                   
        & $\rightarrow$ \textcolor{red}{$Y = 1$} (will  commit delinquency) & \\
\textbf{Else if} &  has anyone (including family members or friends) ever physically attacked you  &\multirow{6}{*}{$\Bigg\} \mathcal{R}_-$} \\
& without a weapon, but you thought you were trying to kill or seriously injure you?   & \\ 
& (Q18B) $\neq$ ``Yes'' \textbf{\emph{and}}  has your friends ever sold hard drugs such as heroin, cocaine  & \\
& and LSD? (Q48AG) $\neq$ ``Yes'' \textbf{\emph{and}}  has your friends ever stolen something worth & \\
&more than \$50? (Q48AH) $\neq$ ``Yes''  \textbf{\emph{and}}  has your friends ever purposely damaged & \\
& or destroyed property that did not belong to them (Q48BA) $\neq$ ``All of them'' & \\                   
        & $\rightarrow$ \textcolor{red}{$Y = -1$} (will not commit delinquency) &                                       \\
\textbf{Else}    &\textcolor{blue}{$Y = \fb(\mathbf{x})$} &  \multicolumn{1}{c}{$f_b$}      \\ \bottomrule                
\end{tabular}
\end{table}

\begin{figure}[h!]
\centering
  \includegraphics[width=0.5\textwidth]{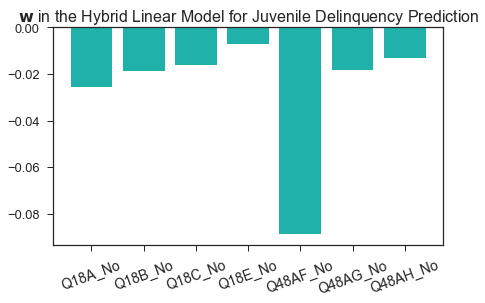}\vspace{-2mm}
\caption{Coefficients in a $\langle f_\text{linear}, \text{Adaboost} \rangle$ model. The model has transparency of 85.5\% and accuracy of 90.6\%. $\theta_+ = 0.81, \theta_- = -0.14$. See Appendix B for the questions in the figure.}\label{fig:juvenile_linear}   \end{figure}

For comparison, we show in Figure \ref{fig:juvenile_linear} the hybrid linear model that substitutes Adaboost, with the same accuracy as the hybrid rule set but larger transparency, 85.5\%. 

The two models are very consistent in choosing the features. Both models include question 18A, 18B, 18C, 48AG and 48AH and there is slight difference in using a couple of other questions.

\subsection{Text Classification}
We apply the hybrid models to two text mining tasks. We convert text datasets into bag-of-words, and each feature is a word.

\paragraph{Review Usefulness Classification}
In this task, the goal is to predict whether a consumer review in Yelp is useful. The dataset we use is a subset of the publically available Yelp Challenge data \footnote{\url{https://www.yelp.com/dataset/challenge}} in 2017. The subset we select contains all the reviews in the original dataset that are about the businesses in the category of ``Doctors''. On Yelp, consumers can vote ``useful'' for a review (but cannot vote ``not useful''). We define a useful review as one that had received at least one useful vote before it was collected into this dataset. With this definition, we obtain 25,901 reviews where 12,970 are not useful, and 12,931 are useful. We randomly partition the data into training (20,721 reviews), and testing (5,180 reviews) sets. We then create the document-term matrix (DTM) in terms of TF-IDF after removing punctuation, numbers, and stopwords and applying stemming. In the DTM, we only use the terms with frequency at least 50. The obtained DTM contains 2,315 columns (terms). We used two different black-box models $f_b$ for this task. The first model is the extreme gradient boosting trees (XGBoost) \citep{chen2016xgboost} with the maximum depth of a tree being 5. The second model is a three-layer fully connected neural network, where the activation function is the sigmoid function, and there are 128 and 16 neurons in the first and second hidden layers. The hyperparameters of the black-box models are selected from a finite grid using hold-out validation.


\paragraph{Movie Success Prediction}
The goal of this task is to predict the financial success of a movie based on its professional critics' reviews. We frame this problem into a binary classification task. In particular, we define as ``Success'' a movie that produced a gross profit of over 30\% above its production budget and define as ``Failure'' otherwise. 
The dataset is publicly available \citep{joshi2010movie} \footnote{\url{www.cs.cmu.edu/~ark/movie$-data/}} and contains reviews on 1,718 movies. We have 616 movies (230 successes and 386 failures) remained after eliminating the movies that did not have the required financial information (production budget and gross profit).  We randomly partition the data into training (492 movies), and testing (124 movies) sets. After merging all reviews on the same movies into one document, we then create the document-term matrix (DTM) with the frequencies of terms after removing punctuation, numbers, and stopwords and applying stemming. In the DTM, we only use the terms with frequency at least 100 and appearing in at least five documents. The obtained DTM contains 2,069 columns (terms). We also use two black-box models for this task. The first model is the XGBoost model using the same hyper-parameters as in the previous task. The second model is the same neural network as in the previous task except that there are 64 neurons in the first hidden layers. The hyperparameters of the black-box models are selected from a finite grid using hold-out validation.

\begin{figure}[h!]
\centering
  \includegraphics[width=0.75\textwidth]{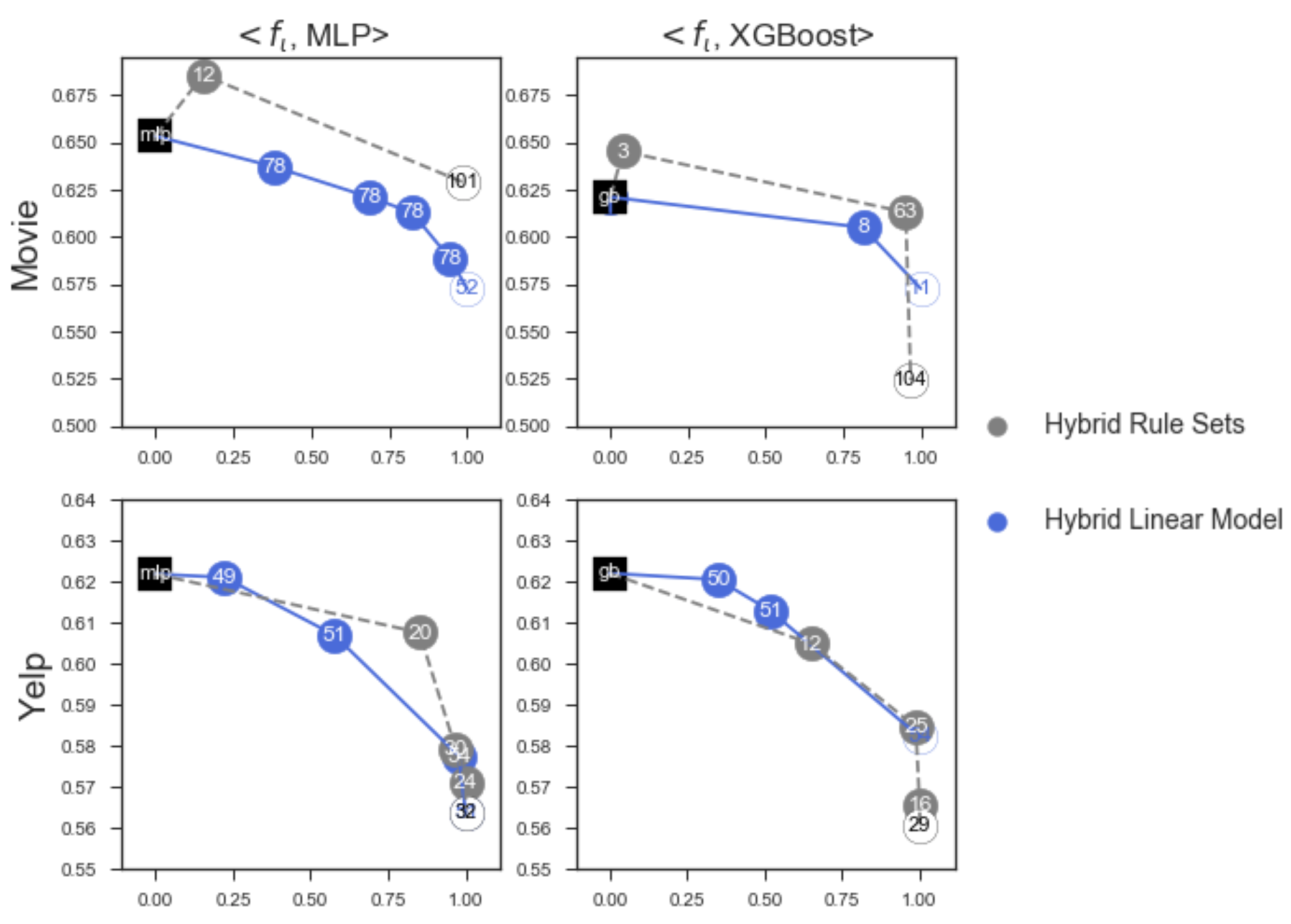}\vspace{-4mm}
\caption{Transparency efficient frontiers of hybrid models. The X-axis represents the transparency and the Y-axis represents the accuracy on the test set. The value inside each circle is the number of items in a hybrid model.}\label{fig:text}   \end{figure}

We train the two types of hybrid models for the tasks above and tune parameters in both methods to obtain a list of models. When training hybrid rule sets, we  binarize the data and consider both the existence and the non-existence of words (considering the negation of features), thus doubling the features.
When training hybrid linear models, we continue to choose $\phi$ to be the hinge loss and $r(x)=\|x\|_1$ and choose the parameters in Algorithm~\ref{alg:APG} as in Section~\ref{sec:struct}. We vary $\alpha_2$ from 0 to 1 to show the trade-off between transparency and accuracy. For each value of $\alpha_2$, we search $\alpha_1$ over a grid using cross-validation to research a good balance between the classification accuracy on the validation set and the number of non-zero coefficeints in $\mathbf{w}$. In particular, we choose $\alpha_1$ such that the number of non-zero coefficeints is no more than 80 in order to obtain an interprable model. The efficient frontiers evaluated on the test set are shown in Figure~\ref{fig:text}. For the Yelp review dataset, the curves for the two models are very similar. They decrease slowly when transparency is small and then more quickly when transparency is larger. For the movie review dataset, it is interesting to observe that the hybrid rule sets models are able to outperform the black-box by a little when transparency is small. This is because rules are less likely to overfit especially the black-box model uses thoursands of features when the rules only uses a very small subset. 

\vspace{-2mm}
\subsubsection*{Case Study: Yelp Review Usefulness Prediction}
\begin{figure}[h!]
\centering
  \includegraphics[width=\textwidth]{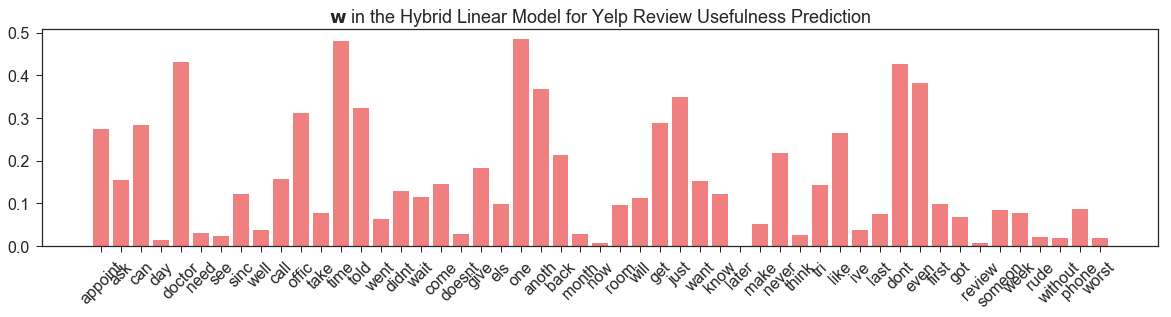}\vspace{-2mm}
\caption{Coefficients in a $\langle f_\text{linear}, \text{XGBoost}\rangle$. The accuracy is 61.3\%, and the transparency is 51.7\%. $\theta_+ = 0.82, \theta_-=0.16.$}\label{fig:yelp_linear}   \end{figure}

We examine hybrid models for the yelp review data set. We choose models that collaborate with XGBoost and with reasonably large transparency. Table \ref{tab:yelp_rs} shows the hybrid rule set model. The positive rule set contains two rules and the negative rule set contains one rule. The rules describe which words need to be included or cannot be included in order to be classified as ``certainly positive'' or ``certainly negative''. Figure \ref{fig:yelp_linear} shows the words selected by the linear model and their corresponding coefficients. Note that the coefficients are all positive in this model, indicating each word listed will contributing to predicting positive (a useful review). However, this does not mean that the linear model only captures the positive class. The threshold for predicting negative is $0.16$, which means if a review does not contain any of the words in Figure \ref{fig:yelp_linear}, or only a few words with low weight, then $\mathbf{w}^\top \mathbf{x}$ will marginally above 0, but less than 0.16 and the instance is predicted as negative. Only when the scores are large enough, for example, containing many keywords from the list, the model will predict positive. If $\mathbf{w}^\top\mathbf{x}$ is between 0.16 and 0.82, the XGBoost model will be activated to make a prediction.

\begin{table}[h]
\centering
\footnotesize
\caption{$\langle f_\text{rule}, \text{XGboost}\rangle$ for juvenile delinquency dataset. The transparency is 64.9\% and the accuracy is 60.5\%, evaluated on the test set.}\label{tab:yelp_rs}
\begin{tabular}{llc}
\toprule
        & \multicolumn{1}{c}{\textbf{Rules}}                                                             & \multicolumn{1}{c}{\textbf{Model}}             \\\hline
\textbf{If} &  the review contains ``wouldnt'' \textbf{\emph{and}} ``ever'' \textbf{\emph{and}} doesn't contain ``left'' &\multirow{3}{*}{$\Bigg\} \mathcal{R}_+$} \\
& \textbf{\emph{OR}} the review contains ``age'' \textbf{\emph{and}} doesn't contain ``worth''&\\
&\textbf{\emph{OR}} the review contains ``let'' \textbf{\emph{and}} doesn't contain ``occas'' or ``lab'' & \\
        & $\rightarrow$ \textcolor{red}{$Y = 1$} (review is useful) & \\ \textbf{Else if} &  the review doesn't contains ``seem'' or ``night'' or ``cant'' or ``saw''  &$ \big\}\mathcal{R}_-$ \\
        & $\rightarrow$ \textcolor{red}{$Y = -1$} (review is unuseful) & \\
\textbf{Else}    &\textcolor{blue}{$Y = \fb(\mathbf{x})$} &  \multicolumn{1}{c}{$f_b$}      \\ \bottomrule                
\end{tabular}\vspace{-2mm}
\end{table}

\vspace{-5mm}
\section{Conclusion and Future Directions}
We proposed a novel framework for learning a Hybrid
Predictive Model that integrates an interpretable model with \textbf{any} black-box model. The interpretable model substitutes the black-box on a subset of data to gain transparency at an efficient cost (sometimes no cost) of predictive performance.  The model is trained to jointly optimize predictive performance, interpretability, and transparency, with carefully designed training algorithms to achieve the best balance among the three.  


The main contribution of this work is that we proposed a novel idea of combining interpretable models with black-box models and proved the collaboration of the two benefits transparency and sometimes even accuracy. Hybrid models provide more choices for users. Instead of choosing between black-box models (transparency equal to zero) and interpretable models (transparency equal to one), hybrid models span the entire spectrum of transparency, and users can choose the best operating point based on the desired transparency and tolerable loss in accuracy. 
Another notable contribution is that our framework is generalizable. In addition to the two types of interpretable substitutes discussed in this paper, our framework can support a variety of interpretable models, such as rule lists (ordered rules), case-based models, decision trees, etc.

Our model can be combined seamlessly with a black-box explainer, working as a pre-step for a black-box explainer: we first construct an interpretable model to capture and explain a sub-region with 100\% fidelity and
leave the rest of the data to a black-box model to predict and an
explainer to provide post hoc explanations.
\subsection*{Future Directions}
For future research, there are several directions worth pursuing. First, we can explore using other forms of interpretable models, such as case-based models:  if an instance is close enough to a prototype, the prediction is determined by the prototype; otherwise, it is sent to a black-box model.  In addition, other regularization terms of interpretable models can be included in the objective function, such as using fewer features or adding a fairness component that penalizes disparity in different subgroups of data.   

In this paper, the black-box is pre-trained, and only the outputs are provided during training. It will be interesting to co-train a black-box and an interpretable model. We hypothesize that the co-training is likely to achieve a better balance among accuracy, interpretability, and transparency, but the computation complexity will increase significantly. It remains a question whether the benefits of co-training are worth the effort.

Finally, the current models only work on structured data and text data. Association rules and linear models are not designed for raw image processing or other problems where the features themselves are not interpretable. The notions of interpretability used in image classification tend to be completely
different from those for structured data where each feature is separately meaningful. There are some recent attempts on building interpretable models for image classification \citep{chen2018looks,Zhang_2018_CVPR} and it will be interesting to combine these models with black-box neural nets such as CNN or ResNet for image classfication.

\subsection*{Appendix A: Proofs for Theorems}
\begin{proof}(of Theorem \ref{theorem:support})
To prove that an optimal model does not contain any rules with support smaller than a threshold, we prove by contradiction that if there's a model that contains such a rule $z$, removing it will always decrease the objective value, thus violating the optimality assumption.

We start with positive rules. For a rule $r \in \mathcal{R}^*_+$, we define
\begin{equation}\label{eqn:theorem31}
    \mathcal{R}^*_{\backslash r} = \{z \in \mathcal{R}^*_+, z\neq r\}.
\end{equation}
We want to find conditions on the support such that the following inequality always holds.
\begin{equation}\label{eqn:i1}
\Lambda(\mathcal{R}^*_{\backslash r})\leq \Lambda(\mathcal{R}^*),
\end{equation}
where
\begin{align}
    \Lambda(\mathcal{R}^*) =& \ell ( \mathcal{R}^*) + \alpha_1 \cdot \Omega(\mathcal{R}^*) - \alpha_2 \cdot \frac{\text{support}(\mathcal{R}^*)}{N} \label{eqn:ieq2} \\
\Lambda(\mathcal{R}^*_{\backslash r}) \leq & \ell ( \mathcal{R}^*_{\backslash r}) + \alpha_1 \cdot \left(\Omega(\mathcal{R}^*) -1\right) - \alpha_2 \cdot \frac{\text{support}(\mathcal{R}^*_{\backslash r})}{N}. \label{eqn:ieq1}
\end{align}
Since we want inequality (\ref{eqn:i1}) to hold for any $r \in \mathcal{R}^*$, we upper bound $\Lambda(\mathcal{R}^*_{\backslash r})$ by upper bounding $\ell (\mathcal{R}^*_{\backslash r})$ and $\text{support}(\mathcal{R}^*_{\backslash r})$.
\begin{equation}\label{eqn:i2}
\ell ( \mathcal{R}^*_{\backslash r}) \leq \ell ( \mathcal{R}^*) + \frac{\text{support}(\mR^*_{\backslash r})}{N},
\end{equation}
with the minimum achieved when instances originally covered by $r$ are all incorrectly classified after removing $r$. 
\begin{equation}\label{eqn:i3}
\text{support}(\mathcal{R}^*_{\backslash r}) \leq \text{support}(\mathcal{R}^*)
\end{equation}
with the minimum achieved when all instances originally covered by $r$ are now covered by $\mR^*_-$, therefore does not change the coverage of $\mR^*$ overall.

Plugging formula (\ref{eqn:i2}) and (\ref{eqn:i3}) into equation (\ref{eqn:ieq1}) and combine it with (\ref{eqn:ieq2}) and (\ref{eqn:i1}) yields
\begin{equation}
    \text{support}(r)\leq N\alpha_1.
\end{equation}
Thus, if $\text{support}(r) \leq N\alpha_1$, removing it from a $\mathcal{R}^*$ will produce a better model. Therefore, such rules do not exist in an optimal model $\mathcal{R}^*$.

Then we follow the similar steps to prove for negative rules. We define $\mathcal{R}^*_{\backslash r}$ as a set of rules where $r$ is removed from $\mR^*_-$. The proofs here use the same steps from inequality (\ref{eqn:theorem31}) to (\ref{eqn:i2}). The effective coverage, however, equals to $\text{support}(\mathcal{R}^*) - \text{support}(\mathcal{R}^*_{\backslash r})$.
    Thus
    \begin{equation}
            \text{support}(r)\leq \frac{N\alpha_1}{1-\alpha_2}.
    \end{equation}
 To summarize, $\mR^*_+$ does not contain any rules with     $\text{support}(r)\leq N\alpha_1$ and $\mR^*_-$ does not contain any rules with $\text{support}(r)\leq \frac{N\alpha_1}{1-\alpha_2}$.
\end{proof}

\begin{proof}(of Theorem \ref{theorem:size})
    We choose the optimal model found till time $t$ to be the benchmark to compare with $\mathcal{R}^*$.
Since $\mathcal{R}^* \in \min \Lambda(\mathcal{R})$,
\begin{equation}\label{eqn:i3}
\lambda^{[t]}\geq \Lambda(\mathcal{R}^*),
\end{equation}
Now we lower bound $\Lambda(\mathcal{R}^*)$, following equation (\ref{eqn:i1})
\begin{equation}\label{eqn:i4}
    \Lambda(\mathcal{R}^*) \geq 0 + \alpha_1 \Omega(\mathcal{R}^*)- \alpha_2 
\end{equation}
Combining inequality (\ref{eqn:i3}) and (\ref{eqn:i4}) yields
\begin{equation}
\Omega(\mathcal{R}^*) \leq \frac{    \lambda^{[t]} + \alpha_2}{\alpha_1}.
\end{equation}
\end{proof}

\begin{proof}(of Theorem \ref{theorem:transparency})
We again compare $\mR^*$ with the best model we found till time $t$ and 
\begin{equation}\label{eqn:i3}
\lambda^{[t]}\geq \Lambda(\mathcal{R}^*),
\end{equation}
Then we lower bound $\Lambda(\mR^*)$,
\begin{equation}
    \Lambda(\mR^*) \geq 0 + \alpha_1\Omega(\mR^*) - \alpha_2 \frac{\text{support}(\mR^*)}{N}
\end{equation}
If $\mR^* \neq \emptyset$, then $\Omega(\mR^*) \geq 1$,  then
\begin{equation}
    \lambda^{[t]} \geq \alpha_1 - \alpha_2\frac{\text{support}(\mR^*)}{N},
\end{equation}
Thus
\begin{equation}
    \text{support}(\mR^*) \geq \frac{N(\alpha_1 - \lambda^{[t]})}{\alpha_2}
\end{equation}

\end{proof}
\subsection*{Appendix B: Questions listed in Figure \ref{fig:juvenile_linear}}
\begin{itemize}
    \item Q18A: has anyone (including family members or friends) ever attacked you with a gun, knife or some other weapon, regardless of when it happened or whether you ever reported to police
    \item Q18B: has anyone (including family members or friends) ever physically attacked you without a weapon, but you thought they were trying to kill or seriously injure you?
    \item Q18C: has anyone (including family members or friends) ever threatened you with a gun or knife, but didn't actually shoot or cut you?
    \item Q18E: has anyone (including family members or friends) ever beat you up with their fists so hard that you were hurt pretty bad?
    \item Q48AF: has your friends ever broken into a vehicle or building to steal something?
    \item Q48AG: has your friends sold hard drugs such as heroin, cocaine, and LSD?
    \item Q48AH: has your friends stolen something worth more than \$50 ?
\end{itemize}
\vskip 0.2in
\bibliography{ms_hybrid,tong,iml,nips_msr,rules_ICDM,jmlr_msr,DRL}
\end{document}